\documentclass[10pt,twocolumn,letterpaper]{article}

\usepackage[pagenumbers]{cvpr} 

\definecolor{cvprblue}{rgb}{0.21,0.49,0.74}
\usepackage[pagebackref,breaklinks,colorlinks,allcolors=cvprblue]{hyperref}

\usepackage[capitalize]{cleveref}
\usepackage{pifont}
\usepackage{booktabs} 
\usepackage{makecell} 
\usepackage{arydshln}

\usepackage[accsupp]{axessibility}  

\def\paperID{4878} 
\def\confName{CVPR}
\def\confYear{2026}

\title{Understanding the Role of Hallucination in Reinforcement Post-Training of Multimodal Reasoning Models}

\author{Gengwei Zhang$^{1}$,\quad Jie Peng$^{2}$,\quad Zhen Tan$^{3}$,\quad Mufan Qiu$^{1}$,\quad Hossein Nourkhiz Mahjoub$^{4}$, \\ Vaishnav Tadiparthi$^{4}$,\quad Kwonjoon Lee$^{4}$,\quad Yanyong Zhang$^{2}$,\quad Tianlong Chen$^{1*}$ \\
$^{1}$ University of North Carolina at Chapel Hill \quad
$^{2}$ University of Science and Technology of China \\
$^{3}$ Arizona State University \quad
$^{4}$ Honda Research Institute, USA \\
{\tt\small *Correspondence to: Tianlong Chen <tianlong@cs.unc.edu>.}
}

\begin{document}

\twocolumn[{
    \renewcommand\twocolumn[1][]{#1}
    \maketitle
    \vspace{-12mm}
    \begin{center}
        \centering\captionsetup{type=figure}
        \includegraphics[width=\linewidth]{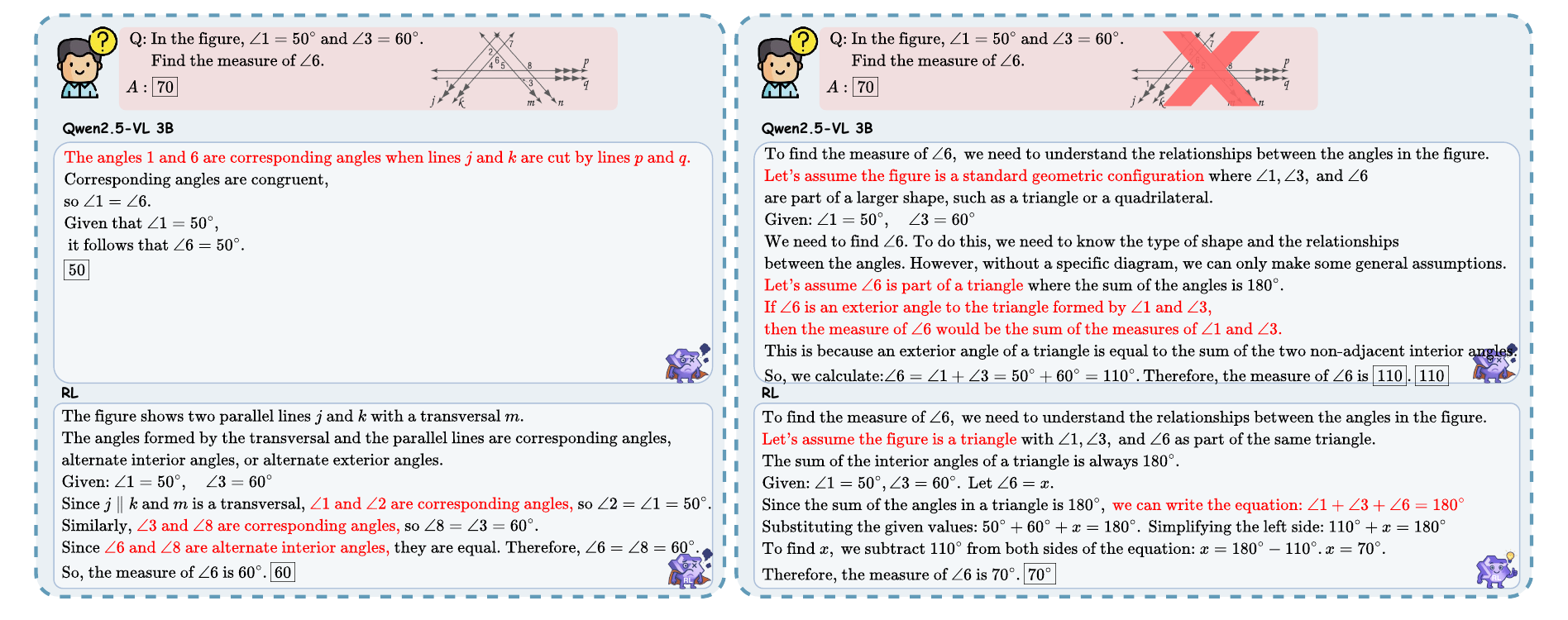}
        \vspace{-8mm}
        \caption{\textit{Case Study.} An example illustrating different hallucination behaviors in multimodal reasoning models. The left side shows the model reasoning with normal visual inputs; in this case, the reinforcement-trained model (bottom-left) produces a noisier reasoning trajectory and ultimately yields an incorrect answer. In contrast, the right side demonstrates that when visual information is removed, the reinforcement-trained model focuses directly on the question itself. With only contextual cues, the model arrives at a correct prediction. Hallucinated contents are marked in \textcolor{red}{red}. \label{fig:teaser}}
        \vspace{-2mm}
    \end{center}
}]


\begin{abstract}
The recent success of reinforcement learning (RL) in large reasoning models has inspired the growing adoption of RL for post-training Multimodal Large Language Models (MLLMs) to enhance their visual reasoning capabilities. Although many studies have reported improved performance, it remains unclear whether RL training truly enables models to learn from visual information. In this work, we propose the \textbf{Hallucination-as-Cue Framework}, an analytical framework designed to investigate the effects of RL-based post-training on multimodal reasoning models from the perspective of model hallucination. Specifically, we introduce hallucination-inductive, modality-specific corruptions that remove or replace essential information required to derive correct answers, thereby forcing the model to reason by hallucination. By applying these corruptions during both training and evaluation, our framework provides a unique perspective for diagnosing RL training dynamics and understanding the intrinsic properties of datasets. Through extensive experiments and analyses across multiple multimodal reasoning benchmarks, we reveal that the role of model hallucination for RL-training is more significant than previously recognized. For instance, we find that RL post-training under purely hallucination-inductive settings can still significantly improve models' reasoning performance, and in some cases even outperform standard training. These findings challenge prevailing assumptions about MLLM reasoning training and motivate the development of more modality-aware RL-based training designs.
\end{abstract}
    
\section{Introduction}
\label{sec:intro}

Recent advances in reasoning-oriented Large Language Models (LLMs)~\cite{jaech2024openai,guo2025deepseek} have demonstrated the effectiveness of reinforcement learning (RL) in encouraging LLMs with emergent reasoning capabilities for solving complex multi-step problems such as mathematical problem solving. Motivated by the advances in textual reasoning, recent studies~\cite{shen2025vlm,huang2025vision} have begun exploring RL-based post-training for \textit{Multimodal Large Language Models} (MLLMs)~\cite{liu2023visual,zhu2023minigpt,wang2024qwen2}. 
Despite the impressive gains in reasoning accuracy reported in recent RL-trained MLLMs~\cite{chen2025sft,shen2025vlm,huang2025vision}, how RL improved the multimodal reasoning capability is still unknown. 

Specifically, given the modality-agnostic nature of final-answer-based reward design, we raise a fundamental question that should be answered before RL-based post-training can be more broadly and reliably extended to more multimodal models: \textit{Does RL-based training truly leverage visual information during training to teach models how to see and reason, or does it primarily strengthen internal reasoning patterns within LLMs?} 
Clarifying this mechanistic question is essential for developing future multimodal reasoning frameworks that achieve genuinely modality-grounded intelligence, with broad applications for domains such as medical imaging diagnosis systems~\cite{lai2025med}.

To answer this critical question, we introduce an analytical framework called \textit{Hallucination-as-Cue} to examine the role of hallucination in RL-based post-training. The core intuition behind this idea is that, as generated reasoning chains become longer, MLLMs tend to rely increasingly on self-generated textual priors rather than grounding their reasoning in visual information, thereby increasing the likelihood of hallucination~\cite{ji2023survey,liu2025more}. Importantly, as illustrated in~\cref{fig:teaser}, hallucination does \textbf{not} always lead to incorrect predictions in visual reasoning. This interesting behavior motivates us to treat hallucination as a diagnostic signal for understanding visual reasoning training.

Specifically, in our \textit{Hallucination-as-Cue} framework, we introduce three modality-specific corruptions that remove or replace essential information required to derive correct answers, and apply them at both training and test time. Because the models are still required to produce responses, this design forces them to generate reasoning steps and answers purely from corrupted inputs and their internally hallucinated reasoning traces. This setup enables a systematic analysis of model behavior during both training and inference, allowing us to study the impact of corruption on training and test data, the effect of model scale, and the properties of existing visual reasoning benchmarks.

As a results, our findings reveal that (1) despite training with corrupted data, RL-based training can still improves performance, and in some cases, matches or even surpasses normal RL training; (2) larger MLLMs benefit even more from hallucinated trajectories; (3) corruption-based inference analysis shows that in some cases visual information can even hinder performance of smaller models. 

In summary, our findings challenge the common assumption that RL post-training can effectively utilize training visual information. Instead, we reveal that current RL-based training would lead models to rely more heavily on textual priors. We hope that our analysis provides new insights into the mechanisms and limitations of RL-based multimodal post-training and inspires future work toward better multimodal reasoning models.

We summarize our contributions as follows:  
(1) We provide the first systematic analysis of RL-based post-training for multimodal reasoning models, examining its effects from the perspective of model hallucination.  
(2) We introduce an innovative framework, \textit{Hallucination-as-Cue}, which enables a diagnostic investigation of the behaviors of training algorithms, reasoning models, and reasoning datasets under hallucination-inductive conditions.
(3) Through extensive and in-depth analyses, we show that the role of hallucination in multimodal reasoning model training is more substantial than previously recognized, motivating future development of more principled RL-based post-training methods for multimodal reasoning models.

\section{Related Work}
\label{sec:related_work}

\subsection{Reasoning in MLLMs}
Early Multimodal Large Language Models (MLLMs), such as the BLIP series~\cite{li2022blip,li2023blip}, primarily focused on vision–language alignment, demonstrating the potential for improved text generation conditioned on visual inputs. LLaVA~\cite{liu2023visual} further advanced multimodal reasoning by introducing the visual instruction tuning framework. Subsequently, more powerful models~\cite{bai2025qwen2,chen2024internvl} have been developed with larger model capacities and training datasets, achieving superior performance on a variety of tasks such as visual question answering (VQA)~\cite{singh2019towards}. However, these models still struggle with complex tasks such as visual mathematical reasoning~\cite{zhang2024mathverse,wang2024measuring}, which demand multi-step and reasoning capabilities. To address this limitation, recent studies have explored various approaches to enhance the reasoning capabilities of MLLMs, including long chain-of-thought generation~\cite{zhang2025improve,thawakar2025llamav}, Monte Carlo tree search~\cite{yao2024mulberry}, and latent-space reasoning~\cite{yang2025machine}. Inspired by the success of reasoning LLMs such as DeepSeek-R1~\cite{guo2025deepseek}, numerous recent works~\cite{yang2025r1,liu2025visual,huang2025vision,shen2025vlm,liu2025noisyrollout,chen2025sft,yang2025visionthink,wang2025sota,wang2025vl} have investigated reinforcement learning (RL)–based post-training strategies for MLLMs through rule-based reward optimization. While the adoption of reinforcement learning has demonstrated notable gains in multimodal reasoning performance, it remains unclear whether these improvements stem from genuine visual understanding of the training images or merely from enhanced text-based reasoning capabilities.

\subsection{Hallucination in Reasoning MLLMs}
Research on hallucination in current reasoning MLLMs is still in its early stage. Most existing studies focus primarily on hallucination mitigation~\cite{wang2025think,wu2025mitigating,zou2024look} or benchmark evaluations~\cite{zhang2024mathverse,dong2025mirage,liu2025more}, where accuracy degradation is treated as an implicit indicator of model hallucination~\cite{zhang2024mathverse,liu2025more}, or rely on larger MLLMs to detect hallucinated outputs~\cite{dong2025mirage}. Most recently, a concurrent work~\cite{asadi2026mirage} shows that MLLMs can retain high accuracy on multimodal benchmarks even without access to images. To the best of our knowledge, no prior work has systematically investigated the impact of hallucination during the training process of reasoning MLLMs.

\section{Preliminaries}
\label{sec:preliminaries}

\noindent
\textbf{Reinforcement Learning for Reasoning Post-training.}
Recently, reinforcement learning (RL) with rule-based reward functions has proven effective in enhancing the reasoning ability of LLMs. These reward functions check whether the reasoning trajectory follows a specific format and whether the predicted answer is correct. Among these methods, the Group Relative Policy Optimization (GRPO) algorithm~\cite{guo2025deepseek} extends standard policy gradient methods by normalizing rewards within a group of sampled completions and thus eliminates the need to train a reward model.

We will use GRPO algorithm as our study case. For a given input prompt $x$, GRPO samples a group of $G$ candidate completions, denoted as $\{y_i\}_{i=1}^G$, from the old policy $\pi_{\theta_{\mathrm{old}}}$. Each completion receives a reward $R_i$, and a group-normalized advantage is computed as
\begin{equation}
    A_i = \frac{R_i - \mu_{\text{group}}}{\sigma_{\text{group}} + \epsilon},
\end{equation}
where $\mu_{\text{group}}$ and $\sigma_{\text{group}}$ are the mean and standard deviation of $\{R_i\}_{i=1}^G$, and $\epsilon$ is a small constant for numerical stability.

The GRPO objective optimizes a PPO-style clipped surrogate~\cite{schulmanProximal2017} with group-relative advantages and a KL penalty, which is expressed as:
\begin{align}
\resizebox{0.89\linewidth}{!}{$
\begin{aligned}
\mathcal{L}_{\text{GRPO}}(\theta)
= &\mathbb{E}\left[
    \frac{1}{G} \sum_{i=1}^G
    \min\left(
        r_i(\theta) A_i,\,
        \text{clip}\left(r_i(\theta), 1 - \epsilon_{\text{clip}}, 1 + \epsilon_{\text{clip}}\right)\,A_i
    \right)
\right]
 \\&- \beta\, D_{\mathrm{KL}}\!\big(\pi_\theta(\cdot \mid x)\,\big\|\,\pi_{\text{ref}}(\cdot \mid x)\big)\\
 \end{aligned}$},
\end{align}
where $r_i(\theta) = \frac{\pi_\theta(y_i \mid x)}{\pi_{\theta_{\mathrm{old}}}(y_i \mid x)}$ and  $\epsilon_{\text{clip}}$ is the clipping threshold, $\pi_{\text{ref}}$ is a frozen reference model, and $\beta$ controls the KL regularization strength. In practice, we take $\pi_{\text{ref}}$ (and the initial $\pi_{\theta_{\mathrm{old}}}$) as the model before GRPO fine-tuning. Specifically, for training MLLMs, the GRPO training is similar to that of LLMs. All modalities are fed into the models in the form of tokens. We simply extend $\pi_\theta(y_i|x)$ to $\pi_{\theta}(y_i|x_t,x_v)$, where $x_t$ is the text prompt and $x_v$ is the vision tokens. 
\section{Hallucination-as-Cue Framework}
\label{sec:method}

In this section, we present our analytical framework, which utilizes hallucination as a diagnostic cue to analyze RL in MLLMs post-training.

\noindent
\textbf{Scope of Study.} Current RL-based MLLMs post-traing methods~\cite{yang2025r1,huang2025vision,shen2025vlm,liu2025noisyrollout,chen2025sft,yang2025visionthink,wang2025sota} are dominated by visual mathematical reasoning tasks with Group Relative Policy Optimization (GRPO)~\cite{guo2025deepseek}. Therefore, following current practice, we main study GRPO method on visual mathematical reasoning~\cite{lu2023mathvista,qiao2025we} or classic visual reasoning tasks~\cite{johnson2017clevr}
In visual mathematical reasoning, each sample typically provides a diagram, such as a geometric figure, along with textual descriptions specifying essential variable conditions, followed by a question for the model to answer. 

\noindent
\textbf{Modality-Specific Corruption.} 
Although hallucination is generally regarded as an undesirable behavior and is commonly associated with performance degradation~\cite{zhang2024mathverse,liu2025more}, hallucinated reasoning can sometimes, by coincidence or due to inherent model bias, produce correct answers. When such cases occur during RL-based post-training, they would be positively reinforced, due to the reward is assigned solely based on the correctness of the final answer. Besides, from an evaluation perspective, the occurrence rate of these false positives can also reveal how a model behaves before and after training, as well as highlight properties of the evaluation datasets.

Therefore, rather than suppressing or mitigating hallucination, we try to leverage it and intentionally encouraging it during both training and testing. Specifically, as shown in~\cref{fig:framework}(a), we design a set of modality-specific corruption strategies that encourage the model to learn or evaluate under the following hallucination-inductive corruptions:

\begin{figure*}
    \centering
    \includegraphics[width=\linewidth]{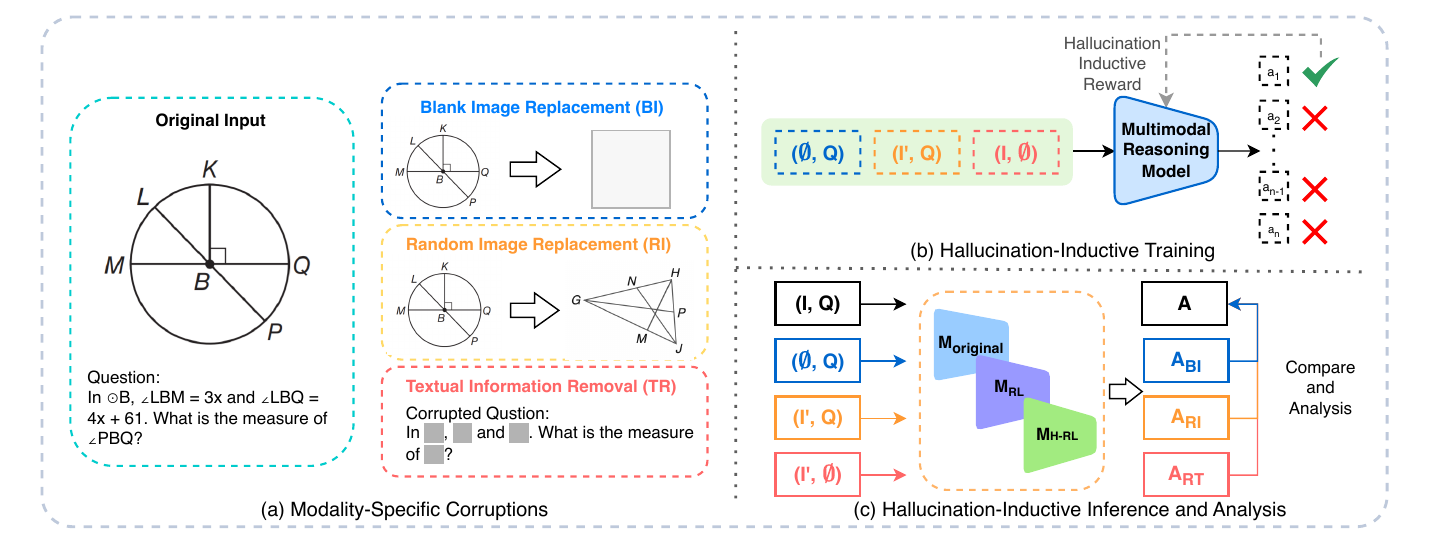}
    \vspace{-8mm}
    \caption{\textit{Hallucination-as-Cue Framework.} \textbf{(a) Modality-Specific Corruptions}: We define three types of data corruptions: \textit{Blank Image Replacement}, \textit{Random Image Replacement}, and \textit{Textual Information Removal}. \textbf{(b) Hallucination-Inductive Training}: We apply these types of modality-specific corruptions to the training data to obtain three models. Since the input information is corrupted, the model learns to hallucinate the corrupted information for inference. We refer to this process as hallucination-inductive training. \textbf{(c) Hallucination-Inductive Inference and Analysis}: We then analyze these three models under the three types of data corruptions to compare task accuracy and examine model behavior.}
    \vspace{-2mm}
\label{fig:framework}
\end{figure*}

\begin{itemize}
    \item \textbf{Blank Image Replacement (BI).} All training images are replaced with blank images, removing all visual information. This forces the model to internally hallucinate the visual content required for reasoning.
    
    \item \textbf{Random Image Replacement (RI).} Each training image is replaced with a randomly selected image from the dataset, creating mismatched text–image pairs. This corruption compels the model to reason and predict answers based on incorrect or irrelevant visual information.
    
    \item \textbf{Textual Information Removal (TR).} All textual conditions and text queries are removed using rule-based matching. When it fails (e.g., < 3 condition matched), we remove the entire text and retain only a template such as ``answer the question step by step''. This corruption forces the model to rely primarily on visual inputs. 
\end{itemize}

\subsection{Hallucination-as-Cue.} 
Building upon the aforementioned corruption types, we introduce the hallucination-inductive training and evaluation strategy, as shown in~\cref{fig:framework}(b) and~\cref{fig:framework}(c). 

\noindent
\textbf{Hallucination-Inductive Training}. During training, RL-based post-training generates $n$ rollout trajectories, each representing a sampled response that the model produces for a given input. When any modality-specific corruption is applied, most of these responses become filled with hallucinated content. Under the rule-based reward scheme used in current RL-based post-training methods, a small subset of these hallucinated responses may still receive positive rewards, which we refer to as \textit{hallucination-inductive rewards}, as illustrated in~\cref{fig:framework}(b). These hallucination-inductive rewards encourage the reasoning model to increase the likelihood of such trajectories and, consequently, to learn reasoning patterns derived from them. In our experiments in~\cref{sec:exp}, we train models on multiple datasets with diverse characteristics~\cite{lu2021inter,leng2025mmr1,johnson2017clevr} to analyze how RL-based post-training learns from hallucination-dominant trajectories by tracking the accuracy trends throughout training and examine how sensitive hallucination-inductive RL training is to different training data.

\noindent
\textbf{Hallucination-Inductive Inference}. At test time, the reasoning model is asked to generate full reasoning chains and produce a final answer. When any modality-specific corruption is applied, accuracy naturally decreases because essential information has been removed. However, the accuracy does not drop to zero, either due to the model’s hallucinated reasoning or simply by chance. As illustrated in~\cref{fig:framework}(c), we compare different models under various corruption settings to better understand their hallucination behaviors.

\section{Experimental Analysis}
\label{sec:exp}

\subsection{Experimental Setup}

\noindent
\textbf{Training Datasets.}
To investigate RL training and model behavior under diverse scenarios, we train models on three datasets: Geometry3K~\cite{lu2021inter}, MMR1-V0~\cite{leng2025mmr1}, and CLEVR~\cite{johnson2017clevr}. Geometry3K is one of the most commonly used datasets for visual mathematical reasoning, containing approximately 3,000 geometry problems with 2,100 training samples, 300 validation samples, and 600 test samples. Unless otherwise specified, we use the Geometry3K training split for model training and its test split for evaluation. MMR1-V0 is a compositional dataset consisting of 7,000 examples covering various types of visual mathematical problems. In addition, we also include CLEVR, a classical visual reasoning dataset, to support our analysis.

\noindent
\textbf{Evaluation Benchmarks.}
Beyond the test split of Geometry3K, we further evaluate our models on four recent multimodal mathematical reasoning benchmarks: MathVision~\cite{wang2024measuring}, MathVerse~\cite{zhang2024mathverse}, MathVista~\cite{lu2023mathvista}, and We-Math~\cite{qiao2025we}. These benchmarks cover different types of visual mathematical problems and vary in difficulty, annotation granularity, and problem distribution, collectively spanning both text-intensive and vision-intensive samples.

\noindent
\textbf{Implementation Details.}
Following prior works~\cite{yang2025r1,liu2025visual,huang2025vision,weifirst}, we use Qwen2.5-VL models~\cite{bai2025qwen2} in our experiments. We use Qwen2.5-VL-3B model for our major investigation and also use Qwen2.5-VL-7B model for study the effect of model size. We follow the implementation of GRPO from the EasyR1 training framework~\cite{zheng2025easyr1} and train the model on each dataset for 15 episodes using the AdamW optimizer with a constant learning rate of $1\times10^{-6}$ and a weight decay of $1\times10^{-2}$. To stabilize training, we apply gradient clipping with a maximum gradient norm of 1.0. For GRPO-specific hyperparameters, we set the rollout size to 5, sampling temperature to 0.7, and KL-divergence weight to 0.01.

\subsection{Analysis Settings}
To provide a comprehensive understanding of RL-based post-training within our proposed framework, we consider the following eight evaluation settings: \\
\textbf{S1:} Standard performance on the training set. \\
\textbf{S2:} Standard performance on the \underline{test set}. \\
\textbf{S3:} Performance of the normally trained model on the corrupted training set. \\
\textbf{S4:} Performance of the normally trained model on the corrupted \underline{test set}. \\
\textbf{S5:} Result of the \textit{hallucination-inductive trained model} on the normal training set. \\
\textbf{S6:} Result of the \textit{hallucination-inductive trained model} on the normal \underline{test set}. \\
\textbf{S7:} Result of the \textit{hallucination-inductive trained model} on the corrupted training set. \\
\textbf{S8:} Result of the \textit{hallucination-inductive trained model} on the corrupted \underline{test set}.

\noindent
S1–S2 serve as baselines that reflect the normal behavior of the initial model or standard RL optimization. S3–S4 provide insight into how a normally trained reasoning model behaves when presented with corrupted inputs. 

When compared with the S1–S4 baselines, S5–S6 reveal how models trained under hallucination-inductive conditions perform on clean data, and S7–S8 indicate whether hallucination-inductive training introduces more correct cases even under corrupted inputs.

\begin{figure}
    \centering
    \includegraphics[width=\linewidth]{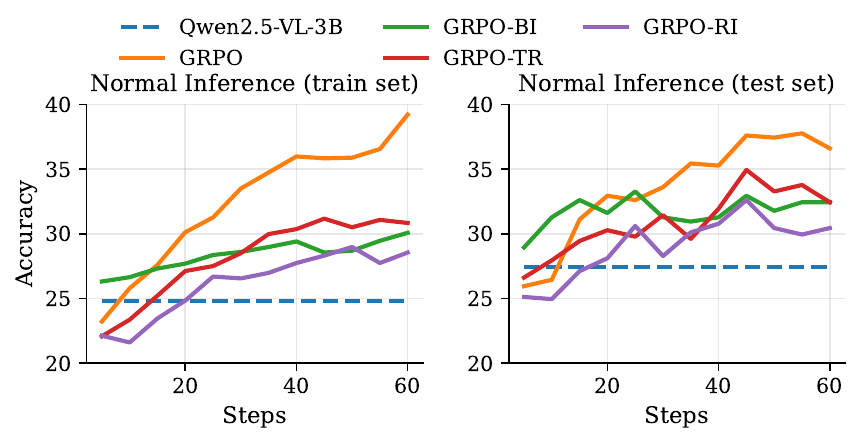}
    \vspace{-7mm}
    \caption{Accuracy of different training regimes on the normal training and test sets.}
    \vspace{-5mm}
\label{fig:norm_inf}
\end{figure}

\begin{figure*}
    \centering
    \includegraphics[width=1.\linewidth]{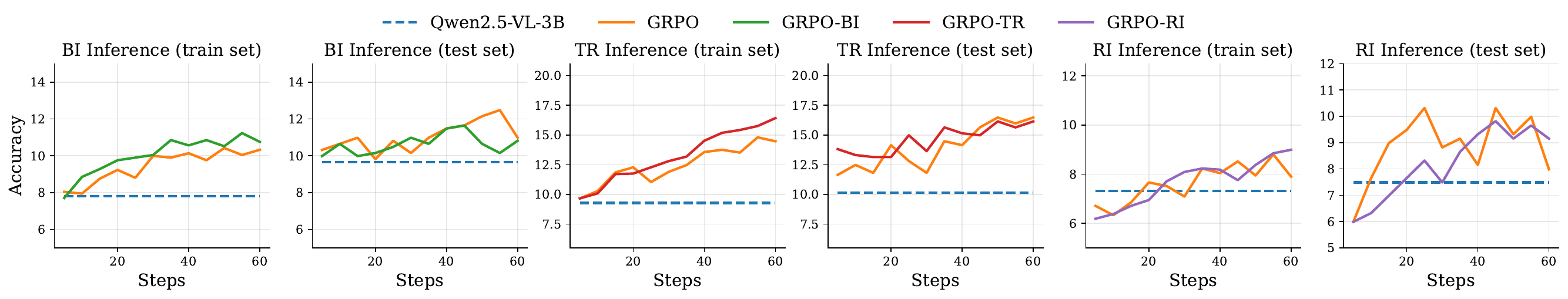}
    \vspace{-7mm}
    \caption{Accuracy of different training regimes on corrupted training and test sets.}
    \vspace{-2mm}
\label{fig:corrupt_inf}
\end{figure*}

\subsection{Hallucination-Inductive RL Training}
\label{sec:hallu-rl-train}
We begin by applying RL training to the Qwen2.5-VL-3B model on the Geometry3K~\cite{lu2021inter} dataset. For each modality-specific corruption, we train a separate model to isolate and observe its effect. Across all training runs, we monitor performance using the S1–S8 settings introduced above. 

As shown in~\cref{fig:norm_inf}, we first analyze model behavior by comparing different corruption settings under S1–S2 and S5–S6, which are evaluated  on original training and test set. With GRPO training, both training and test accuracy are improved over the Qwen-2.5-VL base model. When applying modality-specific corruptions during training, it can be observe that RL-based post-training can still improve model reasoning performance, either on training set and test set. 

Specifically, with \textbf{Blank Image Replacement (BI)}, the model performs “visual’’ reasoning purely based on textual information. Interestingly, it improves both training and test accuracy even in the very early stages of training, indicating that RL-based MLLM training tends to favor text-based signals. The continued performance improvement throughout training further shows that current RL-based multimodal reasoning methods can learn from hallucinated visual reasoning even without access to any actual visual content in the training data. This observation aligns with findings in~\cite{chen2025advancing}, where the authors report that purely textual reasoning data can also improve visual reasoning performance. However, unlike their setting, where text descriptions fully specify the task, in our case the text only provide very narrow problem context, and it is commonly believed that models should not be able to learn effectively from such severely corrupted training data. 

With \textbf{Random Image Replacement (RI)}, performance slightly decreases at the beginning of training, similar to the pattern observed in normal training. However, accuracy increases substantially after a few steps. This suggests that although the training images provide incorrect information, the model does not collapse and can still benefit from RL-based training. Moreover, hallucinated reasoning driven by mismatched visual inputs can also amplify the reasoning capability of MLLMs.

With \textbf{Textual Information Removal (TR)}, the overall trend is similar to that of RI training. Note that although key textual conditions are removed from the prompt during training, a portion of the images may still provides the question context. For example, visual cues such as an arrow or a question mark typically indicate the target to be solved, and many geometry images also provide condition values within the image. In principle, this should allow the model to encounter more ``normal'' cases than in BI and RI training. However, TR training does not show a clear performance gain over BI and RI training, indicating that current RL-based multimodal reasoning training may not yet be effective at leveraging visual information during training.

To better understand this behavior, we revisit the case study in~\cref{fig:teaser} as an illustrative example. From a reasoning perspective, a small portion of positive hallucinated trajectories indeed exhibit correct textual reasoning. Rewarding such trajectories encourages the model to learn effective reasoning behaviors, while negative hallucinated trajectories are discouraged, preventing excessive hallucination.

\begin{table*}[t]
\centering
\small
\caption{Benchmark results comparing different training regimes with Qwen2.5-VL~\cite{bai2025qwen2} baselines. All RL-based training is performed on the Geometry3K~\cite{lu2021inter} dataset. For the 3B model, standard GRPO yields the best average performance. Strikingly, for the 7B model, hallucination-inductive training (GRPO-RI) achieves the highest average score, even outperforming standard GRPO trained on clean data.}
\vspace{-2mm}
\label{tab:geo3k-train}
\begin{tabular}{lcccc|c}
\toprule
\textbf{Model} & \textbf{MathVision (\%)} & \textbf{MathVerse (\%)} & \textbf{MathVista (\%)} & \textbf{WeMath (\%)} & \textbf{AVG (\%)} \\
\midrule
\multicolumn{6}{c}{\textbf{Qwen2.5-VL-3B}} \\
\midrule
Qwen2.5-VL-3B & 18.19 & 34.82 & 51.40 & 54.48 & 39.72 \\
Qwen2.5-VL-3B + GRPO          & 22.73 & 37.72 & 58.40 & 60.11 & \textbf{44.74} \\
Qwen2.5-VL-3B + GRPO-BI       & 20.95 & 35.10 & 56.40 & 56.55 & 42.25 \\
Qwen2.5-VL-3B + GRPO-RI       & 20.86 & 35.76 & 58.00 & 55.17 & 42.45 \\
Qwen2.5-VL-3B + GRPO-TR       & 19.97 & 35.53 & 58.50 & 55.34 & 42.34 \\
\midrule
\multicolumn{6}{c}{\textbf{Qwen2.5-VL-7B}} \\
\midrule
Qwen2.5-VL-7B & 27.70 & 45.20 & 67.00 & 63.68 & 50.89 \\
Qwen2.5-VL-7B + GRPO          & 28.13 & 47.56 & 70.00 & 68.39 & 53.52 \\
Qwen2.5-VL-7B + GRPO-BI       & 28.39 & 48.86 & 68.50 & 66.84 & 53.15 \\
Qwen2.5-VL-7B + GRPO-RI       & 27.27 & 49.90 & 71.40 & 68.33 & \textbf{54.23} \\
Qwen2.5-VL-7B + GRPO-TR       & 26.81 & 49.39 & 70.10 & 67.76 & 53.51 \\
\bottomrule
\end{tabular}
\vspace{-2mm}
\end{table*}

\begin{table*}[t]
\centering
\small
\caption{Benchmark results analyzing the effect of different post-training datasets. We compare the performance of the base Qwen2.5-VL model, the standard GRPO-trained model, and the BI-corrupted GRPO-trained model across three post-training datasets: Geometry3K~\cite{lu2021inter}, MMR1-V0~\cite{leng2025mmr1}, and CLEVR~\cite{johnson2017clevr}.}
\vspace{-2mm}
\label{tab:multi_dataset}
\begin{tabular}{lccccc|c}
\toprule
\textbf{Method} & \textbf{Post-training Dataset} 
& \textbf{MathVision (\%)} & \textbf{MathVerse (\%)} 
& \textbf{MathVista (\%)} & \textbf{WeMath (\%)} & \textbf{AVG (\%)} \\
\midrule
\multicolumn{7}{c}{\textbf{Qwen2.5-VL-3B}}\\
\midrule
Qwen2.5-VL-3B & --          & 18.19 & 34.82 & 51.40 & 54.48 & 39.72 \\
\hdashline
GRPO                  & Geometry3K  & 22.73 & 37.72 & 58.40 & 60.11 & \textbf{44.74} \\
GRPO-BI               & Geometry3K  & 20.95 & 35.10 & 56.40 & 56.55 & 42.25 \\
\midrule
GRPO                  & MMR1-V0     & 26.18 & 39.26 & 65.00 & 62.47 & \textbf{48.23} \\
GRPO-BI               & MMR1-V0     & 24.28 & 40.03 & 61.20 & 61.61 & 46.78 \\
\midrule
GRPO                  & CLEVR       & 23.06 & 35.96 & 58.20 & 55.75 & \textbf{43.24} \\
GRPO-BI               & CLEVR       & 21.51 & 35.05 & 58.20 & 54.20 & 42.24 \\
\midrule
\multicolumn{7}{c}{\textbf{Qwen2.5-VL-7B}}\\
\midrule
Qwen2.5-VL-7B & --          & 27.70 & 45.20 & 67.00 & 63.68 & 50.89 \\
\hdashline
GRPO                  & Geometry3K  & 28.13 & 47.56 & 70.00 & 68.39 & \textbf{53.52} \\
GRPO-BI               & Geometry3K  & 28.39 & 48.86 & 68.50 & 66.84 & 53.15 \\
\midrule
GRPO                  & MMR1-V0     & 30.03 & 50.10 & 70.80 & 70.29 & \textbf{55.31} \\
GRPO-BI               & MMR1-V0     & 27.60 & 48.76 & 73.50 & 69.31 & 54.79 \\
\midrule
GRPO                  & CLEVR       & 26.97 & 48.12 & 67.10 & 65.52 & 51.93 \\
GRPO-BI               & CLEVR       & 26.41 & 47.66 & 68.80 & 66.61 & \textbf{52.37} \\
\bottomrule
\end{tabular}
\end{table*}

\subsection{Hallucination-Inductive Inference}
\label{sec:hallu-rl-infer}
We further analyze model behavior when modality-specific corruptions are applied during inference. As shown in~\cref{fig:corrupt_inf}, although different corruption types lead to varying absolute accuracy levels, both GRPO-trained and hallucination-inductive trained models exhibit clear performance improvements over training steps compared to the baseline Qwen2.5-VL-3B model. This suggests that enhanced reasoning ability increases the likelihood of producing correct answers under corrupted inputs.
Moreover, even when models are trained with the same corruption type, hallucination-inductive training does not demonstrate a clear advantage over standard GRPO training for corrupted inference, on either training or test data. This indicates that the model does not overfit to the hallucinated content during training; instead, it encourages the model to learn general reasoning ability.

\begin{table*}[t]
\centering
\small
\caption{Benchmark results of different models with BI corruption applied to the evaluation data. Performance differences relative to normal inference are indicated in the table.}
\vspace{-2mm}
\label{tab:benchmark_corrupt}
\resizebox{\textwidth}{!}{
\begin{tabular}{lccccc|c}
\toprule
\textbf{Model} & \makecell{\textbf{Inference}\\ \textbf{Corruption?}}
& \textbf{MathVision (\%)} & \textbf{MathVerse (\%)} 
& \textbf{MathVista (\%)} & \textbf{WeMath (\%)} & \textbf{AVG (\%)} \\
\midrule
\multicolumn{7}{c}{\textbf{Qwen2.5-VL-3B}} \\
\midrule
Qwen2.5-VL-3B              
    & no  
    & 18.19 & 34.82 & 51.40 & 54.48 & 39.72 \\
Qwen2.5-VL-3B              
    & yes 
    & 18.91(\textcolor{red}{+0.72})
    & 19.14(\textcolor{ForestGreen}{-15.68})
    & 28.10(\textcolor{ForestGreen}{-23.30})
    & 29.14(\textcolor{ForestGreen}{-25.34})
    & 23.82(\textcolor{ForestGreen}{-15.90}) \\
\hdashline

Qwen2.5-VL-3B+GRPO         
    & no  
    & 22.73 & 37.72 & 58.40 & 60.11 & 44.74 \\
Qwen2.5-VL-3B+GRPO         
    & yes 
    & 20.82(\textcolor{ForestGreen}{-1.91})
    & 21.65(\textcolor{ForestGreen}{-16.07})
    & 28.80(\textcolor{ForestGreen}{-29.60})
    & 33.22(\textcolor{ForestGreen}{-26.89})
    & 26.12(\textcolor{ForestGreen}{-18.62}) \\
\hdashline

Qwen2.5-VL-3B+GRPO-BI      
    & no  
    & 20.95 & 35.10 & 56.40 & 56.55 & 42.25 \\
Qwen2.5-VL-3B+GRPO-BI      
    & yes 
    & 19.80(\textcolor{ForestGreen}{-1.15})
    & 20.23(\textcolor{ForestGreen}{-14.87})
    & 28.50(\textcolor{ForestGreen}{-27.90})
    & 33.28(\textcolor{ForestGreen}{-23.27})
    & 25.45(\textcolor{ForestGreen}{-16.80}) \\
\midrule

\multicolumn{7}{c}{\textbf{Qwen2.5-VL-7B}} \\
\midrule

Qwen2.5-VL-7B              
    & no  
    & 27.70 & 45.20 & 67.00 & 63.68 & 50.89 \\
Qwen2.5-VL-7B              
    & yes 
    & 22.43(\textcolor{ForestGreen}{-5.27})
    & 25.43(\textcolor{ForestGreen}{-19.77})
    & 30.70(\textcolor{ForestGreen}{-36.30})
    & 38.22(\textcolor{ForestGreen}{-25.46})
    & 29.20(\textcolor{ForestGreen}{-21.69}) \\
\hdashline

Qwen2.5-VL-7B+GRPO         
    & no  
    & 28.13 & 47.56 & 70.00 & 68.39 & 53.52 \\
Qwen2.5-VL-7B+GRPO         
    & yes 
    & 22.63(\textcolor{ForestGreen}{-5.50})
    & 25.81(\textcolor{ForestGreen}{-21.75})
    & 33.40(\textcolor{ForestGreen}{-36.60})
    & 40.86(\textcolor{ForestGreen}{-27.53})
    & 30.68(\textcolor{ForestGreen}{-22.84}) \\
\hdashline

Qwen2.5-VL-7B+GRPO-BI      
    & no  
    & 28.39 & 48.86 & 68.50 & 66.84 & 53.15 \\
Qwen2.5-VL-7B+GRPO-BI      
    & yes 
    & 23.26(\textcolor{ForestGreen}{-5.13})
    & 25.41(\textcolor{ForestGreen}{-23.45})
    & 35.70(\textcolor{ForestGreen}{-32.80})
    & 41.78(\textcolor{ForestGreen}{-25.06})
    & 31.54(\textcolor{ForestGreen}{-21.61}) \\
\bottomrule
\end{tabular}
}
\vspace{-4mm}
\end{table*}

\begin{table*}[t]
\centering
\small
\caption{Fine-grained analysis of different visual reasoning problem types using the MathVerse~\cite{zhang2024mathverse} benchmark. BI corruption is applied as the study case for inference corruption.}
\vspace{-2mm}
\label{tab:fingrained}
\begin{tabular}{lccccc}
\toprule
\textbf{Model} 
& \makecell{\textbf{Inference}\\ \textbf{Corruption?}}
& \makecell{\textbf{Text}\\\textbf{Dominant}}
& \makecell{\textbf{Text}\\\textbf{Lite}}
& \makecell{\textbf{Vision}\\\textbf{Intensive}}
& \makecell{\textbf{Vision}\\\textbf{Dominant}}\\
\midrule

Qwen2.5-VL-7B              & no  & 54.82 & 48.73 & 45.94 & 45.81 \\
Qwen2.5-VL-7B              & yes & 50.51 & 28.17 & 20.05 & 23.86 \\
\hdashline
$\Delta$   &   -  & -4.31 & -20.56 & -25.89 & -21.95 \\
\midrule
Qwen2.5-VL-7B + GRPO       & no  & 59.77 & 50.63 & 46.70 & 47.46 \\
Qwen2.5-VL-7B + GRPO       & yes & 52.28 & 30.58 & 21.57 & 24.49 \\
\hdashline
$\Delta$   &  -   & -7.49 & -20.05 & -25.13 & -22.97 \\
\midrule
Qwen2.5-VL-7B + GRPO-BI    & no  & 58.76 & 48.98 & 45.18 & 47.21 \\
Qwen2.5-VL-7B + GRPO-BI    & yes & 52.28 & 27.92 & 20.56 & 26.02 \\
\hdashline
$\Delta$  &  -   & -6.48 & -21.06 & -24.62 & -21.19 \\
\bottomrule
\end{tabular}
\vspace{-2mm}
\end{table*}

\subsection{Benchmark Results and Model Scaling.}
Besides accuracy on Geometry3K dataset, in~\cref{tab:geo3k-train}, we report evaluation results other visual reasoning benchmarks. Consistent with our findings on the Geometry3K dataset, apply modality-specific corruptions during RL training still show positive impact to the multimodal reasoning model when compared with the baseline Qwen2.5-VL model.

With the Qwen2.5-VL-3B model, all hallucination-inductive trained variants show improvements across all benchmark evaluations, with particularly significant gains on MathVista, where GRPO-RI and GRPO-TR achieve performance comparable to normal GRPO training. Although the average performance across benchmarks is similar under different modality-specific corruptions, applying each corruption type in training yields different degrees of improvement on different evaluation benchmarks.

\noindent
\ding{226} \textbf{Larger Models Can Learn Stronger Reasoning from Hallucinated Inputs.}
Furthermore, when studying the Qwen2.5-VL-7B model, we observe that all RL training variants with modality-specific corruptions perform on par with standard training that uses correctly paired image--text inputs. More surprisingly, the RI training variant even \textbf{surpasses} standard training in terms of average performance.

This finding challenges the current understanding of multimodal reasoning training. We emphasize that RL-based training may not effectively utilize visual inputs as commonly expected. One possible interpretation is that GRPO-like training does not fully rely on multimodal grounding signals; instead, it primarily amplifies the inherent reasoning capabilities of the underlying language model. Moreover, larger models exhibit a stronger ability to generate and learn from hallucinated positive trajectories, suggesting that model capacity plays a critical role in benefiting from such training signals. Specifically, the 7B model improves from 9.7\% to 14.1\% on Geometry3K-BI, while the 3B model increases from 7.6\% to 10.4\%.

\subsection{Effect of Post-training Datasets} Given the surprising results above, we further investigate whether these observations remain consistent across different post-training datasets. The results are presented in~\cref{tab:multi_dataset}. For simplicity, we apply BI corruption across all datasets and train both Qwen2.5-VL-3B and Qwen2.5-VL-7B models to further validate our conclusions.

Consistent with our findings on Geometry3K, hallucination-inductive training on compositional mathematical datasets such as MMR1-V0, and on traditional visual reasoning datasets such as CLEVR, also yields significant improvements over the baseline. This indicates that the effect of hallucination-inductive training is not limited to a specific training distribution, but instead generalizes across different types of training datasets. Moreover, as model size increases, hallucination-inductive training achieves performance that is on par with, or even superior to, standard GRPO training on visual reasoning tasks.

\subsection{Corruption during Benchmark Evaluation}
Our framework also provide a way for us to analyze the properties of current evaluation benchmarks.
Given the variety of evaluation benchmarks, we examine how models behave when corruption is applied during evaluation. The results are presented in~\cref{tab:benchmark_corrupt}. We apply the BI corruption to all benchmarks and compare performance against standard (uncorrupted) inference. As expected, model performance drops substantially in most cases, though the magnitude of degradation varies across benchmarks.

Interestingly, when evaluating the Qwen2.5-VL-3B model on the MathVision benchmark, BI-corrupted inference yields even higher accuracy than normal inference. This is likely because MathVision contains challenging problems sourced from real math competitions, where visual information can sometimes act as a distraction rather than a helpful cue. In such case, for smaller or non-reasoning-oriented models such as Qwen2.5-VL-3B, the presence or absence of the image may therefore make little difference, and in some cases, removing the image can even simplify the reasoning process. This phenomenon disappears when using larger models, such as Qwen2.5-VL-7B, or when applying RL-based post-training, both of which strengthen the model’s reasoning ability and make visual input more informative rather than distracting.

Comparing normal training and BI-corrupted training under BI-corrupted inference, the average performance differences are similar, which aligns with the analysis in~\cref{sec:hallu-rl-infer}. However, the detailed effects vary across specific benchmarks and model configurations. For instance, on the MathVista benchmark, the performance drop of BI-corrupted training is much smaller than that with normal GRPO training. This can be attributed to the nature of the problem types and the intrinsic difficulty of the benchmark.

Therefore, we further leverage the fine-grained question-type annotations in MathVerse~\cite{zhang2024mathverse} to study how RL-based post-training and input corruptions influence model performance across different categories of questions. Specifically, we separately evaluate models on \textit{Text Dominant}, \textit{Text Lite}, \textit{Vision Intensive}, and \textit{Vision Dominant} questions, where the amount of textual information decreases from the former to the latter (more details are provided in the Appendix and in~\cite{zhang2024mathverse}). The results are presented in~\cref{tab:fingrained}. We adopt BI corruption as the default corruption setting.

Using the Qwen2.5-VL-7B model, we observe that performance on \textit{Text Dominant} questions drops the least under BI-corrupted inference, which is expected since current multimodal model rely heavily on textual cues. With RL-based training, the largest performance gains also occur in the \textit{Text Dominant} category. Moreover, we find that BI-corrupted training results in smaller performance degradation than normal RL post-training on both \textit{Vision Dominant} and \textit{Text Dominant} questions under using BI-corrupted inference, which suggests that BI training may encourages the model to rely less excessively on whichever modality is dominant, leading to more balanced reasoning behavior. 

\subsection{Discussion}
We provide further discussion of the findings revealed by our Hallucination-as-Cue framework, with additional analysis presented in the appendix.

\textbf{Why does corrupted training still improve performance?}
First, corrupted training does not corrupt the model, as most hallucinated trajectories are negative and are ignored or discouraged during training. Moreover, as discussed in~\cite{shenfeld2025rl}, RL training is much more resilient to forgetting~\cite{kirkpatrick2017overcoming,zhang2023slca} than supervised fintuning (SFT), and therefore does not shift the base model aggressively.

Furthermore, using the case study in~\cref{fig:teaser} as an example, from a reasoning perspective, a small portion of positive hallucinated trajectories do exhibit correct textual reasoning patterns. Rewarding such trajectories encourages the model to learn effective reasoning behaviors.

Additionally, as shown in~\cref{sec:hallu-rl-infer}, hallucination-inductive training does not lead to higher hallucination-inductive testing accuracy compared to standard training, indicating that the model learns general reasoning capability rather than simply mimicking hallucinated content during training.

\noindent
\textbf{Why do larger models show more improvements?}
First, larger models have a greater capacity to generate positive hallucinated trajectories. Concretely, the 7B model has an initial accuracy of 9.7\% on Geometry3K-BI, compared to 7.6\% for the 3B model. Moreover, larger models are better able to learn from such trajectories, increasing accuracy to 14.1\% (a 4.4\% improvement), compared to 10.4\% (a 2.8\% improvement) for the 3B model, on Geometry3K-BI.

\section{Conclusion}
\label{sec:conclusion}
In this work, we present the \textit{Hallucination-as-Cue} framework, a systematic approach for analyzing RL-based post-training in multimodal reasoning through the lens of hallucination. By introducing three modality-specific corruptions during both training and inference, our framework reveals how reinforcement learning operates on multimodal reasoning data and how models behave under hallucination-inductive conditions. Our study yields the following key findings: (1) Hallucination-inductive training, despite operating on severely corrupted data, consistently improves performance and, in some cases, matches or even surpasses standard RL training. (2) Larger MLLMs benefit more from hallucinated trajectories, suggesting that RL primarily amplifies the inherent reasoning capacity of the underlying LLM. (3) Hallucination-inductive inference analysis shows that visual information can even hinder the performance of smaller models. 
In summary, our findings challenge the common assumption that RL post-training can effectively utilize visual information. We hope that our analysis provides new insights into the limitations of RL-based multimodal training and inspires future improvements

\textbf{Limitations.} Although we provide an initial analysis, the underlying mechanisms driving the observed behaviors are complex and multifaceted, and require further investigation. Additionally, our study focuses primarily on mainstream RL-based training, while other reasoning paradigms, such as latent-space reasoning, are left for future work.


\section*{Acknowledgement}
This project is partially supported by the grants from Honda Research
Institute USA.

{
    \small
    \bibliographystyle{ieeenat_fullname}
    \bibliography{main}
}


\end{document}